\documentclass[cameraready]{Interspeech}

\title{Post-Training Speech Enhancement Language Models\\with Perceptual Rewards}
\author[affiliation={1}]{Frédéric}{Berdoz}
\author[affiliation={1}]{Luca A.}{Lanzendörfer}
\author[affiliation={1}]{Antonis}{Asonitis}
\author[affiliation={1}]{Roger}{Wattenhofer}
\address{
    $^1$ ETH Zürich, Switzerland
}
\email{\{fberdoz, lucala, aasonitis, wattenhofer\}@ethz.ch}


\newcommand{\gain}[1]{\rlap{\,{\scriptsize\textcolor{green!50!black}{(+#1)}}}}
\usepackage[capitalise]{cleveref}
\usepackage{multirow}
\usepackage{threeparttable}
\usepackage{stfloats}

\usepackage{tabularx}
\usepackage{array}
\usepackage{subcaption}
\newcolumntype{Y}{>{\centering\arraybackslash}X}
\newcolumntype{Z}[1]{>{\centering\arraybackslash}p{#1}}

\usetikzlibrary{arrows.meta, decorations.pathreplacing, positioning}
\usepackage{pgfplots}
\pgfplotsset{compat=1.18}

\begin{document}
\maketitle
\begin{abstract}
Speech enhancement language models achieve strong results when trained on discrete audio tokens, but their optimization relies on token-level cross-entropy rather than the perceptual metrics used for evaluation. We introduce a post-training stage for autoregressive speech enhancement language models using Group Sequence Policy Optimization (GSPO) with multi-metric perceptual rewards. Our method directly optimizes non-differentiable quality metrics (DNSMOS, WER, and UTMOS) as reward signals, without learned surrogates or offline preference pairs. Applied to two autoregressive base models, UniSE and GenSE, our approach achieves state-of-the-art results on the DNS2020 benchmark. A human evaluation ablation further shows that the composite multi-metric reward is preferred over any single-metric variant, confirming that multi-reward optimization avoids the reward hacking observed with single-metric training.
\end{abstract}

\section{Introduction}
\label{sec:intro}

Speech enhancement (SE) aims to recover clean speech from degraded signals affected by noise, reverberation, bandwidth limitation, or packet loss. While traditional approaches based on time-domain or time-frequency-domain models~\cite{luo2019convtasnet, defossez2020real, hu2020dccrn} have been effective for specific distortion types, recent work reframes SE as a sequence-to-sequence problem using autoregressive language models operating on discrete audio tokens~\cite{wang2024selm, kang2025llase, li2024masksr, yan2025unise, yao2025gense}. These language model-based approaches achieve state-of-the-art results by building on large-scale pretraining and the generalization of autoregressive transformers. However, they are all trained exclusively with supervised cross-entropy loss on token sequences.

In natural language processing, the training pipeline for large language models (LLMs) has converged on three stages: pretraining on large corpora, supervised fine-tuning (SFT) on curated data, and post-training with reinforcement learning (RL)~\cite{ouyang2022training, rafailov2023direct, shao2024deepseekmath}. The post-training stage is critical, aligning the model with evaluation criteria that the supervised loss does not directly capture. The key insight is that next-token prediction is a proxy for what we actually care about, and RL bridges this gap by directly optimizing the true objective~\cite{guo2025deepseek}. This three-stage pipeline has not yet been completed for autoregressive speech enhancement language models, which stop at the SFT stage.

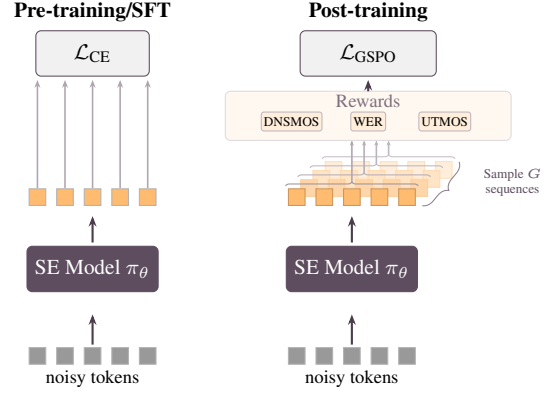
\begin{figure}[!t]
    \centering
    \definecolor{dpurple}{HTML}{5E4C5F}
\definecolor{mgray}{HTML}{999999}
\definecolor{figold}{HTML}{FFBB6F}

\begin{tikzpicture}[xscale=1.35,
    box/.style={draw=dpurple, rounded corners=2pt, minimum height=0.6cm, minimum width=1.4cm,
                font=\footnotesize, align=center, fill=#1},
    box/.default=dpurple!8,
    sq/.style={draw=black!30, fill=#1, minimum size=2.2mm, inner sep=0pt},
    arr/.style={-{Stealth[length=4pt]}, thick, dpurple!80!black},
    lbl/.style={font=\scriptsize, align=center},
    metricbox/.style={draw=dpurple!40, fill=figold!25, rounded corners=1pt,
                      font=\tiny, inner sep=1pt, minimum height=0.25cm},
]

\node[lbl, font=\footnotesize\bfseries] at (2.0, 4.55) {Pre-training/SFT};

\foreach \i in {0,...,4}
    \node[sq=mgray] at (1.46 + \i*0.27, 0) {};
\node[lbl] at (2.0, -0.3) {noisy tokens};

\draw[arr] (2.0, 0.15) -- (2.0, 0.55);

\node[box=dpurple, text=white] (lm1) at (2.0, 1.15) {SE Model $\pi_\theta$};

\draw[arr] (2.0, 1.5) -- (2.0, 1.85);

\foreach \i in {0,...,4}
    \node[sq=figold] at (1.46 + \i*0.27, 2.1) {};

\foreach \i in {0,...,4}
    \draw[-{Stealth[length=3pt]}, dpurple!50, semithick]
        (1.46 + \i*0.27, 2.22) -- (1.46 + \i*0.27, 3.65);

\node[box, minimum width=1.5cm] (ce) at (2.0, 4.0) {$\mathcal{L}_{\text{CE}}$};

\node[lbl, font=\footnotesize\bfseries] at (4.7, 4.55) {Post-training};

\foreach \i in {0,...,4}
    \node[sq=mgray] at (4.0 + \i*0.27, 0) {};
\node[lbl] at (4.54, -0.3) {noisy tokens};

\draw[arr] (4.54, 0.15) -- (4.54, 0.55);

\node[box=dpurple, text=white] (lm2) at (4.54, 1.15) {SE Model $\pi_\theta$};

\draw[arr] (4.54, 1.5) -- (4.54, 1.85);

\foreach \i in {0,...,4}
    \node[sq=figold!25, draw=figold!45] at (4.0+\i*0.27 + 0.36, 2.1 + 0.36) {};
\draw[decorate, decoration={brace, amplitude=1.5pt, raise=0.5pt}, dpurple!35]
    (4.25, 2.59) -- (5.55, 2.59);

\foreach \i in {0,...,4}
    \node[sq=figold!40, draw=figold!55] at (4.0+\i*0.27 + 0.24, 2.1 + 0.24) {};
\draw[decorate, decoration={brace, amplitude=1.5pt, raise=0.5pt}, dpurple!40]
    (4.13, 2.47) -- (5.43, 2.47);

\foreach \i in {0,...,4}
    \node[sq=figold!60, draw=figold!70] at (4.0+\i*0.27 + 0.12, 2.1 + 0.12) {};
\draw[decorate, decoration={brace, amplitude=1.5pt, raise=0.5pt}, dpurple!45]
    (4.01, 2.35) -- (5.31, 2.35);

\foreach \i in {0,...,4}
    \node[sq=figold, draw=figold!80!black] at (4.0 + \i*0.27, 2.1) {};
\draw[decorate, decoration={brace, amplitude=1.5pt, raise=0.5pt}, dpurple!60]
    (3.89, 2.23) -- (5.19, 2.23);

\draw[decorate, decoration={brace, amplitude=4pt, raise=1pt}, dpurple!70]
    (5.57, 2.59) -- (5.21, 2.01);
\node[dpurple, font=\tiny, anchor=west, align=center] at (5.75, 2.3) {Sample $G$\\sequences};

\draw[-{Stealth[length=2pt]}, dpurple!40, semithick] (4.90, 2.66) -- (4.90, 2.85);
\draw[-{Stealth[length=2pt]}, dpurple!42, semithick] (4.78, 2.54) -- (4.78, 2.85);
\draw[-{Stealth[length=2pt]}, dpurple!45, semithick] (4.66, 2.42) -- (4.66, 2.85);
\draw[-{Stealth[length=2pt]}, dpurple!50, semithick] (4.54, 2.30) -- (4.54, 2.85);

\fill[figold!10, draw=dpurple!20, rounded corners=2pt]
    (3.3, 2.85) rectangle (6.1, 3.5);
\node[font=\scriptsize, dpurple!80] at (4.7, 3.38) {Rewards};
\node[metricbox] at (3.95, 3.1) {DNSMOS};
\node[metricbox] at (4.70, 3.1) {WER};
\node[metricbox] at (5.45, 3.1) {UTMOS};

\draw[arr] (4.7, 3.5) -- (4.7, 3.65);

\node[box, minimum width=1.8cm] (loss) at (4.7, 4.0) {$\mathcal{L}_{\text{GSPO}}$};

\end{tikzpicture}
    \caption{Overview of our approach. Left: the base SE model is trained with cross-entropy, which provides per-token loss. Right: GSPO post-training samples $G$ outputs per input and scores each complete sequence with perceptual metrics (DNSMOS, WER, UTMOS), providing per-sequence reward that directly optimizes output quality.}
    \label{fig:overview}
\end{figure}

The train-eval gap in SE is clear: models trained with cross-entropy loss are evaluated using perceptual quality metrics such as DNSMOS~\cite{reddy2021dnsmos}, WER~\cite{radford2023robust}, and UTMOS~\cite{saeki2022utmos}. These evaluation metrics are fundamentally different from cross-entropy, since lower cross-entropy does not guarantee higher DNSMOS or lower WER. Prior attempts to bridge this gap have relied on learned surrogates, such as MetricGAN~\cite{fu2019learning, fu2021metricgan} which trains a discriminator to approximate PESQ, and GSEPF~\cite{li2025aligning} which constructs offline preference pairs using UTMOS as a proxy. However, single-metric optimization is susceptible to reward hacking, a phenomenon where the model learns to exploit artifacts that inflate the targeted score while degrading other quality dimensions~\cite{deoliveira2024pesqetarian}. This motivates a multi-metric reward approach.
\begin{figure}[t]
    \centering
    \includegraphics[width=\columnwidth]{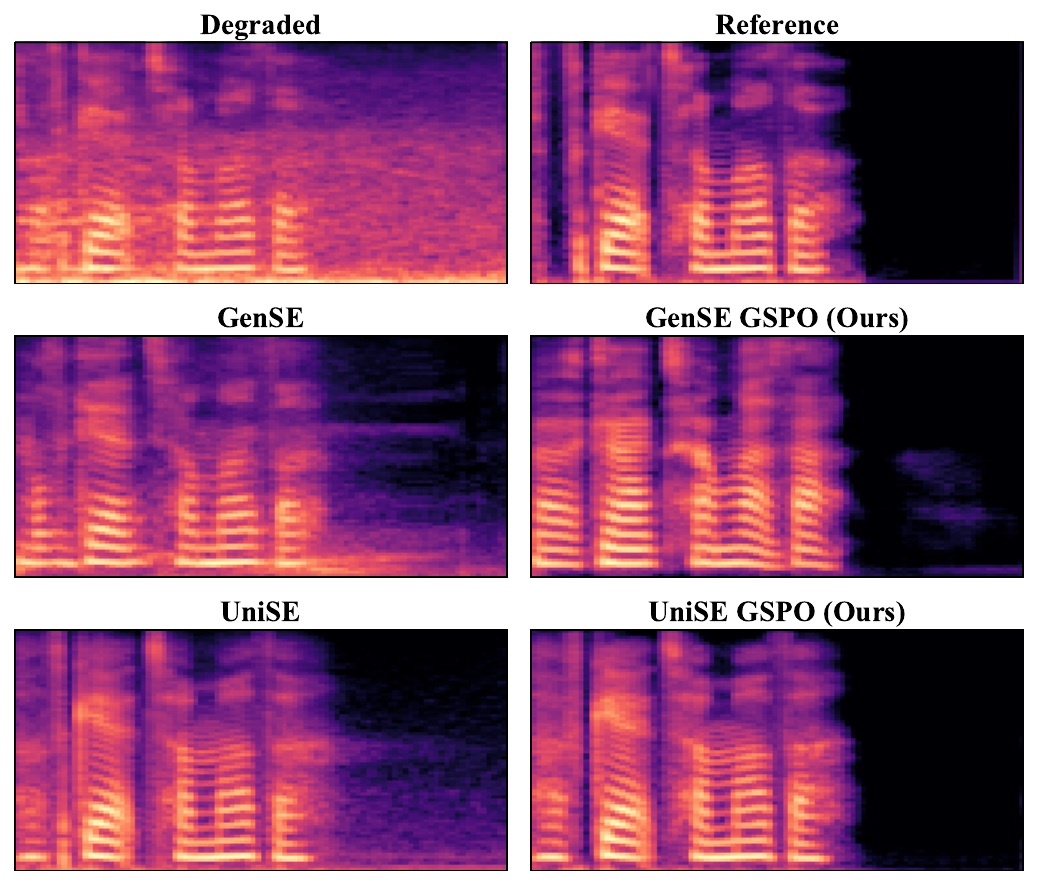}
    \caption{Qualitative comparison of speech enhancement performance after post-training with composite rewards applied to existing autoregressive SE models.}
    \label{fig:mel_spec_comparison}
\end{figure}
To this end, we propose to complete the training pipeline for SE language models by adding a post-training stage using Group Sequence Policy Optimization (GSPO)~\cite{zheng2025gspo}, as illustrated in \cref{fig:overview}. GSPO samples multiple outputs per input, scores them with a reward function, computes group-relative advantages, and applies a clipped policy gradient update at the sequence level, all without a learned critic or value network. Unlike token-level methods such as GRPO~\cite{shao2024deepseekmath}, GSPO defines the importance ratio over full sequence likelihoods, improving training stability. This makes GSPO particularly suitable for SE, where the reward signals (DNSMOS, WER, UTMOS) are non-differentiable but cheap to compute. Unlike MetricGAN or GSEPF, our approach directly uses the actual evaluation metrics as online reward signals, with no learned surrogates or offline data construction.

\medskip
\noindent Our contributions can be summarized as follows:
\begin{itemize}
    \item We introduce post-training for autoregressive speech enhancement language models, applying GSPO with multi-metric perceptual rewards and completing the pretrain-SFT-RL pipeline established in NLP.
    \item We design a composite reward combining DNSMOS, WER, and UTMOS, and show through a human evaluation ablation that the composite reward is preferred over any single-metric variant, confirming that multi-reward training avoids reward hacking.
    \item We achieve state-of-the-art results on the DNS2020 and DNS5 benchmarks by applying GSPO post-training to two autoregressive base models, UniSE~\cite{yan2025unise} and GenSE~\cite{yao2025gense}.
\end{itemize}
\section{Related Work}
\label{sec:related}

\smallskip
\noindent\textbf{LM-Based Speech Enhancement.}
Recent work has framed SE as a token prediction task using autoregressive language models. SELM~\cite{wang2024selm} tokenizes speech into discrete SSL tokens and uses a language model for contextual enhancement. LLaSE-G1~\cite{kang2025llase} uses a LLaMA backbone unifying multiple enhancement tasks. MaskSR~\cite{li2024masksr} extends masked generative modeling to full-band 44.1 kHz restoration, while AnyEnhance~\cite{zhang2025anyenhance} introduces a self-critic mechanism for iterative quality refinement. UniSE~\cite{yan2025unise} provides a unified decoder-only autoregressive framework for speech restoration, extraction, and separation. GenSE~\cite{yao2025gense} uses hierarchical two-stage generation with semantic and acoustic tokens. These methods are trained with supervised objectives and do not incorporate a post-training optimization stage.

\smallskip
\noindent\textbf{Metric Optimization for Speech Enhancement.}
Several approaches directly optimize SE models for perceptual metrics. MetricGAN~\cite{fu2019learning} and MetricGAN+~\cite{fu2021metricgan} train a GAN discriminator as a surrogate for PESQ, enabling gradient-based optimization of a non-differentiable metric. GSEPF~\cite{li2025aligning} applies offline DPO~\cite{rafailov2023direct} to SE using UTMOS as a proxy for human preferences. However, these approaches either rely on learned surrogates (introducing approximation error and training instability) or require offline preference pair construction. The PESQetarian~\cite{deoliveira2024pesqetarian} further shows that single-metric optimization can degrade other quality dimensions, which motivates multi-metric reward design.

\smallskip
\noindent\textbf{Preference Optimization for Speech Generation.}
Preference optimization has been applied to speech generation, primarily for text-to-speech. SpeechAlign~\cite{zhang2024speechalign} applies iterative DPO to codec language models, DLPO~\cite{chen2025dlpo} integrates RL with diffusion model losses, Koel-TTS~\cite{hussain2025koeltts} uses ASR-guided and speaker-verification-guided preferences, and FPO~\cite{yao2025finegrained} proposes token-level preference annotations. Concurrent work FlowSE-GRPO~\cite{wang2026flowsegrpo} applies GRPO to a flow-matching SE model using DNSMOS and speaker similarity as rewards. Our work differs in targeting autoregressive discrete-token SE language models, using GSPO~\cite{zheng2025gspo} for sequence-level optimization, and providing a human evaluation ablation that validates multi-metric reward composition.

\section{Methodology}
\label{sec:method}

\subsection{Base Models}
\label{sec:base_models}

We apply GSPO post-training to two autoregressive SE language models. \textbf{UniSE}~\cite{yan2025unise} is a decoder-only autoregressive LM that unifies speech restoration, speaker extraction, and separation using neural audio codec tokens at 16 kHz. \textbf{GenSE}~\cite{yao2025gense} is an autoregressive LM that uses hierarchical two-stage generation (first semantic tokens, then acoustic tokens) with a single-quantizer neural codec and token chain prompting for timbre preservation. Both models follow the same autoregressive paradigm as text LLMs, generating clean token sequences left-to-right with causal attention, and are trained with a cross-entropy loss. We initialize GSPO from the public SFT checkpoints of each model.

\subsection{GSPO for Speech Enhancement}
\label{sec:gspo}

Given a noisy input $\mathbf{x}$, the policy $\pi_\theta$ (initialized from a base model) produces enhanced speech by autoregressively generating a discrete token sequence $\mathbf{y}$. For each training input, we sample $G$ complete outputs $\{\mathbf{y}^1, \ldots, \mathbf{y}^G\}$ from $\pi_\theta$ using standard sequential decoding with temperature. Each sampled output is decoded to a waveform and scored using a reward function $R(\mathbf{x}, \mathbf{y})$.

\begin{table*}[ht]
    \centering
    \small
    \caption{Comparison of speech enhancement models on the DNS2020 blind test set using DNSMOS P.835 metrics (SIG, BAK, OVRL) across three conditions: synthetic with reverb (150 files), synthetic without reverb (150 files), and real recordings (300 files). GSPO denotes models post-trained with Group Sequence Policy Optimization using a DNSMOS+UTMOS+WER composite reward. Best results per column are in \textbf{bold}. GSPO improves every metric for every base model (\textcolor{green!50!black}{gains} shown in brackets).}
    \label{tab:dns2020_dnsmos}
    \renewcommand{\arraystretch}{0.9}
    \setlength{\tabcolsep}{1pt}
    \begin{tabularx}{\linewidth}{p{2.3cm} YYYYYYYYY}
    \toprule
     & \multicolumn{3}{c}{With Reverb} & \multicolumn{3}{c}{No Reverb} & \multicolumn{3}{c}{Real Recordings} \\
    \cmidrule(lr){2-4} \cmidrule(lr){5-7} \cmidrule(lr){8-10}
    Model & SIG & BAK & OVRL & SIG & BAK & OVRL & SIG & BAK & OVRL \\
    \midrule
    Noisy & 2.03 & 1.65 & 1.53 & 3.50 & 2.81 & 2.62 & 3.16 & 2.68 & 2.36 \\
    \midrule
    Conv-TasNet~\cite{luo2019convtasnet} & 2.42 & 2.71 & 2.01 & 3.09 & 3.34 & 3.00 & 3.10 & 2.98 & 2.41 \\
    Demucs~\cite{defossez2020real} & 2.51 & 2.64 & 2.22 & 3.12 & 3.26 & 3.01 & 2.97 & 2.87 & 2.29 \\
    Inter-SubNet~\cite{chen2023inter} & 2.65 & 2.58 & 2.36 & 3.46 & 3.82 & 3.10 & 3.26 & 3.57 & 2.81 \\
    CDiffuSE~\cite{lu2022conditional} & 2.54 & 2.30 & 2.19 & 3.29 & 3.64 & 3.05 & 3.20 & 3.10 & 2.78 \\
    SGMSE~\cite{richter2023speech} & 2.73 & 2.74 & 2.43 & 3.50 & 3.71 & 3.14 & 3.30 & 2.89 & 2.79 \\
    StoRM~\cite{lemercier2023storm} & 2.95 & 3.14 & 2.52 & 3.51 & 3.94 & 3.21 & 3.41 & 3.38 & 2.94 \\
    SELM~\cite{wang2024selm} & 3.16 & 3.58 & 2.70 & 3.51 & 4.10 & 3.26 & 3.59 & 3.44 & 3.12 \\
    Voicefixer~\cite{liu2022voicefixer} & 3.43 & 4.02 & 3.13 & 3.50 & 4.11 & 3.25 & 3.29 & 3.96 & 2.99 \\
    MaskSR~\cite{li2024masksr} & 3.53 & 4.07 & 3.25 & 3.59 & 4.12 & 3.34 & 3.33 & 4.04 & 3.06 \\
    AnyEnhance~\cite{zhang2025anyenhance} & 3.50 & 4.04 & 3.20 & 3.64 & 4.18 & 3.42 & 3.49 & 3.98 & 3.16 \\
    LLaSE-G1~\cite{kang2025llase} & 3.59 & 4.10 & 3.33 & 3.66 & 4.17 & 3.42 & 3.57 & 4.07 & 3.29 \\
    \midrule
    UniSE~\cite{yan2025unise} & 3.68 & 4.16 & 3.43 & 3.66 & 4.16 & 3.43 & 3.57 & 4.02 & 3.26 \\
    UniSE + GSPO & 3.71\gain{.03} & 4.20\gain{.04} & 3.49\gain{.06} & 3.70\gain{.04} & 4.19\gain{.03} & 3.48\gain{.05} & \textbf{3.63}\gain{.06} & \textbf{4.13}\gain{.11} & \textbf{3.37}\gain{.11} \\
    GenSE~\cite{yao2025gense} & 3.51 & 3.95 & 3.16 & 3.64 & 4.17 & 3.41 & 3.10 & 3.58 & 2.60 \\
    GenSE + GSPO & \textbf{3.76}\gain{.25} & \textbf{4.21}\gain{.26} & \textbf{3.53}\gain{.37} & \textbf{3.75}\gain{.11} & \textbf{4.23}\gain{.06} & \textbf{3.55}\gain{.14} & 3.55\gain{.45} & 4.04\gain{.46} & 3.22\gain{.62} \\
    \bottomrule
    \end{tabularx}
    \end{table*}

GSPO~\cite{zheng2025gspo} builds on GRPO~\cite{shao2024deepseekmath} but operates at the sequence level rather than the token level. It computes a group-relative advantage for each sample by normalizing rewards within the group:
\begin{equation}
\label{eq:advantage}
    A^j_i = \frac{R(\mathbf{x}_i, \mathbf{y}^j_i) - \mu_i}{\sigma_i},
\end{equation}
where $\mu_i$ and $\sigma_i$ are the mean and standard deviation of rewards across the $G$ samples for input $\mathbf{x}_i$. The key difference from GRPO is that GSPO defines the importance ratio over the full sequence likelihood rather than per-token ratios, and applies sequence-level clipping:
\begin{equation}
\label{eq:gspo}
    \mathcal{L} = \mathbb{E}\left[\min\left(\rho \cdot A, \text{clip}(\rho, 1{-}\epsilon, 1{+}\epsilon) \cdot A\right)\right] - \beta D_\text{KL}(\pi_\theta \Vert \pi_\text{ref}),
\end{equation}
where $\rho = \pi_\theta(\mathbf{y} \mid \mathbf{x}) / \pi_\text{old}(\mathbf{y} \mid \mathbf{x})$ is the sequence-level importance ratio. This sequence-level formulation improves training stability compared to token-level GRPO. Like GRPO, GSPO eliminates the critic network required by PPO~\cite{schulman2017proximal}, reducing memory requirements.

\subsection{Reward Function Design}
\label{sec:reward}

We design a composite reward combining three metrics that capture complementary aspects of speech quality:
\begin{equation}
\label{eq:reward}
    R(\mathbf{x}, \mathbf{y}) =  \text{DNSMOS} + (1 - \text{WER}) +   \text{UTMOS},
\end{equation}
where DNSMOS~\cite{reddy2021dnsmos} is a non-intrusive MOS predictor for overall speech quality, WER is computed using Whisper-Large-V3~\cite{radford2023robust} to measure content preservation (subtracted from 1 so that higher is better), and UTMOS~\cite{saeki2022utmos} is a neural MOS predictor capturing naturalness. The three rewards are weighted equally.

This multi-metric formulation is motivated by the observation that single-metric optimization can lead to reward hacking~\cite{deoliveira2024pesqetarian}, where optimizing DNSMOS alone may produce outputs that score high on the targeted metric while simultaneously degrading other quality dimensions. By combining complementary metrics (perceptual quality, intelligibility, and naturalness), we aim to create a more robust reward signal. In \cref{sec:ablation}, we validate this design with a human evaluation showing that listeners prefer the composite reward over any single-metric variant.

\subsection{Training Details}
\label{sec:training_details}

We train on 20k paired noisy-clean samples of 5-second clips at 16 kHz, sourced from the DNS-Challenge datasets, and synthesized for different noise types, such as reverberation, natural noise and static noise. For each input, we sample $G = 4$ outputs. We use AdamW with a learning rate of $1 \times 10^{-5}$, $\beta_1 = 0.9$, $\beta_2 = 0.999$, and weight decay $0.01$. Training runs for 3 epochs of 1000 steps each (3000 total) with a 100-step linear warmup, a batch size of 2 with gradient accumulation over 4 steps (effective batch size 8), clipping $\epsilon = 0.2$, max gradient norm 1.0, temperature 1.0, and fp16 mixed precision. Each model is trained on a single NVIDIA RTX 6000 in 10 hours (single-metric) to 22 hours (composite). At inference, the post-trained model generates a single output with the same cost as the base model.

\begin{table*}[ht]
    \centering
    \caption{Comparison of speech enhancement systems on the DNS5 blind test set using personalized DNSMOS (pDNSMOS) P.835 metrics (pSIG, pBAK, pOVRL) across two tracks: Track 1 Headset (389 files) and Track 2 Speakerphone (364 files). GSPO denotes models post-trained with Group Sequence Policy Optimization using a DNSMOS+UTMOS+WER composite reward. Best results per column are in \textbf{bold}. GSPO improves every metric for every base model (\textcolor{green!50!black}{gains} shown in brackets).}
    \label{tab:dns5_pdnsmos}
    \small
    \renewcommand{\arraystretch}{0.9}
    \begin{tabularx}{\linewidth}{p{2.5cm} YYYYYY}
    \toprule
     & \multicolumn{3}{c}{Track 1: Headset} & \multicolumn{3}{c}{Track 2: Speakerphone} \\
    \cmidrule(lr){2-4} \cmidrule(lr){5-7}
    Model & pSIG & pBAK & pOVRL & pSIG & pBAK & pOVRL \\
    \midrule
    Noisy & 4.15 & 2.37 & 2.71 & 4.05 & 2.16 & 2.50 \\
    \midrule
    TEA-PSE~\cite{ju2023tea} 3.0 & 4.12 & 4.05 & 3.65 & 3.99 & 3.95 & 3.49 \\
    NAPSE~\cite{yan2023npu} & 3.81 & 3.99 & 3.38 & 3.92 & 4.17 & 3.56 \\
    UniFlow$_\text{DDPM}$~\cite{wang2025uniflow} & 4.24 & 3.99 & 3.73 & 4.09 & 3.88 & 3.56 \\
    UniFlow$_\text{FM}$~\cite{wang2025uniflow} & 4.20 & 4.01 & 3.70 & 4.06 & 3.89 & 3.54 \\
    UniFlow$_\text{MF}$~\cite{wang2025uniflow} & 4.18 & 3.99 & 3.67 & 4.04 & 3.87 & 3.51 \\
    LLaSE-G1~\cite{kang2025llase} & 4.21 & 3.99 & 3.72 & 4.08 & 3.84 & 3.55 \\
    \midrule
    GenSE~\cite{yao2025gense} & 4.13 & 3.61 & 3.41 & 3.92 & 3.07 & 2.96 \\
    GenSE + GSPO & 4.65\gain{.52} & 4.61\gain{1.00} & 4.45\gain{1.04} & 4.61\gain{.69} & 4.53\gain{1.46} & 4.36\gain{1.40} \\
    UniSE~\cite{yan2025unise} & 4.52 & 4.43 & 4.21 & 4.47 & 4.41 & 4.15 \\
    UniSE + GSPO & \textbf{4.75}\gain{.23} & \textbf{4.76}\gain{.33} & \textbf{4.63}\gain{.42} & \textbf{4.73}\gain{.26} & \textbf{4.76}\gain{.35} & \textbf{4.61}\gain{.46} \\
    \bottomrule
    \end{tabularx}
    \end{table*}
\section{Experiments}
\label{sec:experiments}

We evaluate GSPO post-training on the DNS2020 blind test set and conduct a human evaluation ablation on the reward composition.

\subsection{Setup}
\label{sec:setup}

We conduct an ablation using a human evaluation to understand the effect of various reward signals on the overall model performance. To this end, we conduct a pairwise preference test with 21 raters on UniSE, comparing five conditions: the SFT baseline (no RL), GSPO with DNSMOS reward only, GSPO with UTMOS reward only, GSPO with WER reward only, and GSPO with the composite reward. Participants are instructed to use wired headphones and be in a quiet environment. The test is conducted using an online tool.\footnote{https://www.mabyduck.com/}

We evaluate on the DNS2020 blind test set~\cite{reddy2020interspeech} across three conditions: synthetic with reverb (150 files), synthetic without reverb (150 files), and real recordings (300 files). Objective quality is measured using DNSMOS P.835~\cite{reddy2021dnsmos} (SIG, BAK, OVRL). We additionally evaluate on the DNS5 blind test set~\cite{dubey2024icassp} across two tracks: Track~1 Headset (389 files) and Track~2 Speakerphone (364 files), using personalized DNSMOS (pDNSMOS) P.835 metrics (pSIG, pBAK, pOVRL). We compare UniSE and GenSE, both with and without GSPO post-training using the composite reward, against a range of baselines including traditional SE models (Conv-TasNet~\cite{luo2019convtasnet}, Demucs~\cite{defossez2020real}, Inter-SubNet~\cite{chen2023inter}), diffusion-based models (CDiffuSE~\cite{lu2022conditional}, SGMSE~\cite{richter2023speech}, StoRM~\cite{lemercier2023storm}), and LM-based models (SELM~\cite{wang2024selm}, VoiceFixer~\cite{liu2022voicefixer}, MaskSR~\cite{li2024masksr}, AnyEnhance~\cite{zhang2025anyenhance}, LLaSE-G1~\cite{kang2025llase}). For the DNS5 evaluation, we additionally compare against UniFlow~\cite{wang2025uniflow}, TEA-PSE 3.0~\cite{ju2023tea} and NAPSE~\cite{yan2023npu}. We present a qualitative comparison in \cref{fig:mel_spec_comparison} to illustrate the impact of post-training on speech enhancement quality.

\subsection{Human Evaluation: Reward Ablation}
\label{sec:ablation}

\begin{table}[t]
\centering
\caption{Post-training ablation study using Human evaluation on UniSE conditions measured with head-to-head win rates (\%) and Bradley-Terry Elo ratings. The composite (Comp.) reward is consistently the top-ranked variant, while single-metric DNSMOS optimization exhibits reward hacking, degrading perceived quality below the SFT baseline (Base.).}
\label{tab:human_eval}

\begin{tabularx}{\linewidth}{p{1cm}Z{0.7cm}YZ{0.9cm}YZ{1.1cm}Y}
\toprule
 & \textbf{Comp.} & \textbf{WER} & \textbf{UTMOS} & \textbf{Base.} & \textbf{DNSMOS} & \textbf{Elo} \\
\midrule
Comp. & --- & 52.5 & 57.1 & 71.4 & 100.0 & 1571 \\
WER & 47.5 & --- & 61.0 & 40.0 & 50.0 & 1541 \\
UTMOS & 42.9 & 39.0 & --- & 51.7 & 100.0 & 1494 \\
Baseline & 28.6 & 60.0 & 48.3 & --- & 66.7 & 1476 \\
DNSMOS& 0.0 & 50.0 & 0.0 & 33.3 & --- & 1335 \\
\bottomrule
\end{tabularx}
\end{table}

To validate the multi-metric reward design, we conduct a pairwise human evaluation comparing UniSE with different GSPO reward configurations against the SFT baseline.  Each participant is shown a degraded signal and two random options for model generations. No ties are allowed. A total of 10 random samples from a test set created using clean speech from HiFiTTS-2~\cite{langman2025hifitts} and noise sourced from DEMAND~\cite{thiemann2013diverse} and RIR NOISES~\cite{ko2017study} are shown to each participant.
\cref{tab:human_eval} shows the preference rate of each condition when compared head-to-head.
The composite reward is clearly preferred over the baseline (71\% win rate), while single-metric variants show mixed results. DNSMOS-only optimization is preferred less than the baseline (37\% win rate), indicating that optimizing a single perceptual metric can degrade overall perceived quality, a likely case of reward hacking. UTMOS-only achieves 55\% and WER-only 40\% against the baseline.
Additionally, we compute Elo ratings from all pairwise matchups using the Bradley-Terry model. The composite reward ranks first (Elo 1571), closely followed by WER (1541), while DNSMOS ranks last (1335), well below even the baseline (1476). This confirms that multi-metric optimization produces outputs that listeners genuinely prefer across all comparisons, while single-metric training, particularly on DNSMOS, actively degrades perceived quality.
We find the asymmetry between metrics to be revealing: DNSMOS-only optimization degrades quality below the baseline (Elo 1335 vs.\ 1476), while UTMOS-only remains near parity (Elo 1494). This difference likely stems from what each metric captures. DNSMOS measures noise suppression characteristics that can be superficially inflated, whereas UTMOS is trained on human MOS annotations and is harder to exploit without genuinely improving naturalness. The composite reward avoids these failure modes by requiring simultaneous improvement across complementary quality dimensions.

\subsection{DNS2020 Results}
\label{sec:dns}

\cref{tab:dns2020_dnsmos} presents the main results. Adding GSPO post-training consistently improves all DNSMOS metrics for both base models on the DNS2020 blind test set. GenSE + GSPO achieves the best results on synthetic conditions (OVRL 3.53 with reverb, 3.55 without reverb), while UniSE + GSPO achieves the best results on real recordings (OVRL 3.37). Both GSPO-enhanced models outperform all baselines, including recent LM-based methods such as AnyEnhance~\cite{zhang2025anyenhance} and LLaSE-G1~\cite{kang2025llase}.

\subsection{DNS5 Results}
\label{sec:dns5}

\cref{tab:dns5_pdnsmos} reports results on the DNS5 blind test set using personalized DNSMOS metrics. GSPO post-training yields substantial improvements, with gains even larger than on DNS2020. GenSE + GSPO improves pOVRL from 3.41 to 4.45 on Track~1 (+1.04) and from 2.96 to 4.36 on Track~2 (+1.40), surpassing all baselines by a wide margin. UniSE + GSPO achieves the best overall scores on Track~1 (pOVRL 4.63). The larger gains on DNS5 compared to DNS2020 likely reflect the lower baseline performance of GenSE on these conditions, leaving more room for improvement through post-training. These results also show that GSPO post-training generalizes to personalized evaluation metrics (pDNSMOS) that are not in the reward function.

\section{Conclusion}
\label{sec:conclusion}

We introduce post-training for autoregressive speech enhancement language models using GSPO with multi-metric perceptual rewards. By directly optimizing non-differentiable quality metrics (DNSMOS, WER, UTMOS) as reward signals, our approach closes the train-eval gap without learned surrogates. Applied to UniSE and GenSE, GSPO post-training achieves state-of-the-art results on the DNS2020 benchmark. Our human evaluation ablation confirms that the composite reward is preferred over any single-metric variant, showing that multi-reward optimization avoids the reward hacking observed when training on individual metrics. Post-training complements architecture and data scaling as a third axis for improving SE language models.

\section{Generative AI Use Disclosure}
Generative AI tools were used to assist with polishing the manuscript, writing code for the experiments, and beautifying figures. All outputs were reviewed and verified by the authors, who take full responsibility for the content of this paper.

\bibliographystyle{IEEEtran}
\bibliography{references}
\end{document}